\pdfoutput=1

\documentclass[11pt]{article}

\usepackage{acl}

\usepackage{times}
\usepackage{latexsym}

\usepackage[T1]{fontenc}

\usepackage[utf8]{inputenc}

\usepackage{microtype}

\usepackage{inconsolata}

\usepackage{graphicx}
\usepackage{algpseudocode}
\usepackage{algorithm}
\usepackage{amsmath}
\usepackage{amsfonts}
\usepackage{arydshln}

\title{UniBridge: A Unified Approach to Cross-Lingual Transfer Learning for Low-Resource Languages}
\author{
    Trinh Pham$^{1*}$,
    Khoi M. Le$^{2}\thanks{$\quad$ Equal Contribution.}$, 
    Luu Anh Tuan$^{3}\thanks{Corresponding author.}$
    \\
    $^{1}$Ho Chi Minh City University of Technology (HCMUT), VNU-HCM, Ho Chi Minh City, Vietnam \\
    $^{2}$VinAI Research, Vietnam \\
    $^{3}$Nanyang Technological University, Singapore \\
    \texttt{phkhanhtrinh23@gmail.com, v.khoilm1@vinai.io, anhtuan.luu@ntu.edu.sg}
}

\begin{document}
\maketitle
\begin{abstract}
In this paper, we introduce UniBridge (Cross-Lingual Transfer Learning with Optimized Embeddings and Vocabulary), a comprehensive approach developed to improve the effectiveness of Cross-Lingual Transfer Learning, particularly in languages with limited resources. Our approach tackles two essential elements of a language model: the initialization of embeddings and the optimal vocabulary size. Specifically, we propose a novel embedding initialization method that leverages both lexical and semantic alignment for a language. In addition, we present a method for systematically searching for the optimal vocabulary size, ensuring a balance between model complexity and linguistic coverage. Our experiments across multilingual datasets show that our approach greatly improves the F1-Score in several languages. UniBridge is a robust and adaptable solution for cross-lingual systems in various languages, highlighting the significance of initializing embeddings and choosing the right vocabulary size in cross-lingual environments.
\end{abstract}

\section{Introduction}

Recently, multilingual pre-trained language models (LMs) have significantly advanced natural language processing (NLP) tasks, narrowing the performance gap between English and various other languages. Multilingual pre-trained models such as XLM-R \citep{conneau-etal-2020-unsupervised} and mBERT \citep{devlin-etal-2019-bert} are currently strong models for effectively cross-lingual transfer \citep{hu2020xtreme, artetxe-etal-2020-cross, le-etal-2024-lampat}. However, these models pose a limitation that they are pre-trained on a limited set of approximately 100 languages, leaving a substantial void for the vast array of the world's nearly 7000 languages \citep{van-esch-etal-2022-writing}. The resultant disparity disproportionately affects low-resource languages that are not covered in their pre-trained corpora \citep{wu-dredze-2020-languages, pfeiffer-etal-2020-mad}, impeding their performance compared to their high-resource counterparts.

\begin{figure}[ht]
    \centering
    \includegraphics[width=\linewidth]{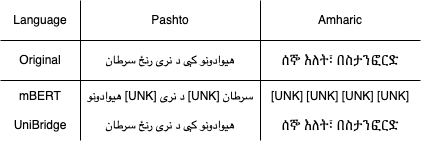}
    \caption{Some languages/scripts are not covered in the pre-trained corpora. Hence, the pre-trained tokenizer will eventually produce many unknown tokens which corrupts the sentence's meaning and results in poor performance.}
    \label{fig:unseen_script}
\end{figure}
\vspace{-0.4cm}

Recent efforts propose the use of adapters to mitigate the knowledge gap in low-resource languages prior to transferring knowledge for specific tasks \citep{pfeiffer-etal-2020-mad, ustun-etal-2020-udapter, ansell-etal-2021-mad-g}. These methods adapt the pre-trained LMs to a new language by utilizing monolingual data, enabling the model to acquire a robust representation of the target language before receiving knowledge from the source language. Despite enhanced performance in languages not included in the pre-trained corpora, these approaches still exhibit poor performance in languages with unseen scripts (i.e., the scripts that are not presented in the pre-training corpora; see Figure \ref{fig:unseen_script}). To address the issue of unseen scripts, existing studies \citep{artetxe-etal-2020-cross, pfeiffer-etal-2021-unks} propose acquiring a new vocabulary embedding for newly discovered languages. However, these methods heavily rely on manually configuring the vocabulary size and initializing the embedding matrix.

Furthermore, recent Cross-Lingual Transfer Learning studies focus on English due to its abundant pre-trained data and impressive task performance, our experiments reveal that high performance in English tasks does not necessarily guarantee successful transfer to other languages, particularly low-resource languages. Therefore, we suggest an automated method utilizing the LMs to identify the most suitable set of source languages for knowledge aggregation, leading to notable performance improvements over single-source language transfer.

Our research empirically tested the effectiveness of newly random initialized embeddings and fixed vocabulary size. We then introduce an efficient technique for determining the optimal vocabulary size for new languages, utilizing the syntactic and semantic insights from the pre-trained LMs. In addition, we present an innovative method for transferring knowledge from multiple sources, which allows the model to choose the best combination of source languages to improve the overall performance. Our results contribute to the ongoing discussion about managing linguistic diversity in NLP, particularly for languages with limited resources, emphasizing the importance of a detailed and inclusive strategy in creating multilingual pre-trained LMs.

We evaluate our approach on sequence tagging tasks (e.g. NER, POS) and classification (e.g. NLI) with two strong baselines, mBERT and XLM-R, and observe a significant increase in the F1 and accuracy score \footnote{\url{https://github.com/VinAIResearch/UniBridge}}. In summary, our contributions are:
\begin{itemize}
    \item We propose a novel approach to automatic search for a suitable vocabulary size to adapt to a new language.
    \vspace{-3mm}
    \item We propose a new strategy to initialize the embedding that leverages the syntactic and semantic knowledge encoded in the pre-trained LMs to address the missing tokens when adapting to low-resource languages.
    \vspace{-3mm}
    \item We propose a method to aggregate multi-source transfer learning to enhance the performance on cross-lingual transfer tasks. We show that multi-source can outperform effective multi-language learning.
\end{itemize}

\section{Methodology}

\begin{figure*}
    \centering
    \includegraphics[width=\linewidth]{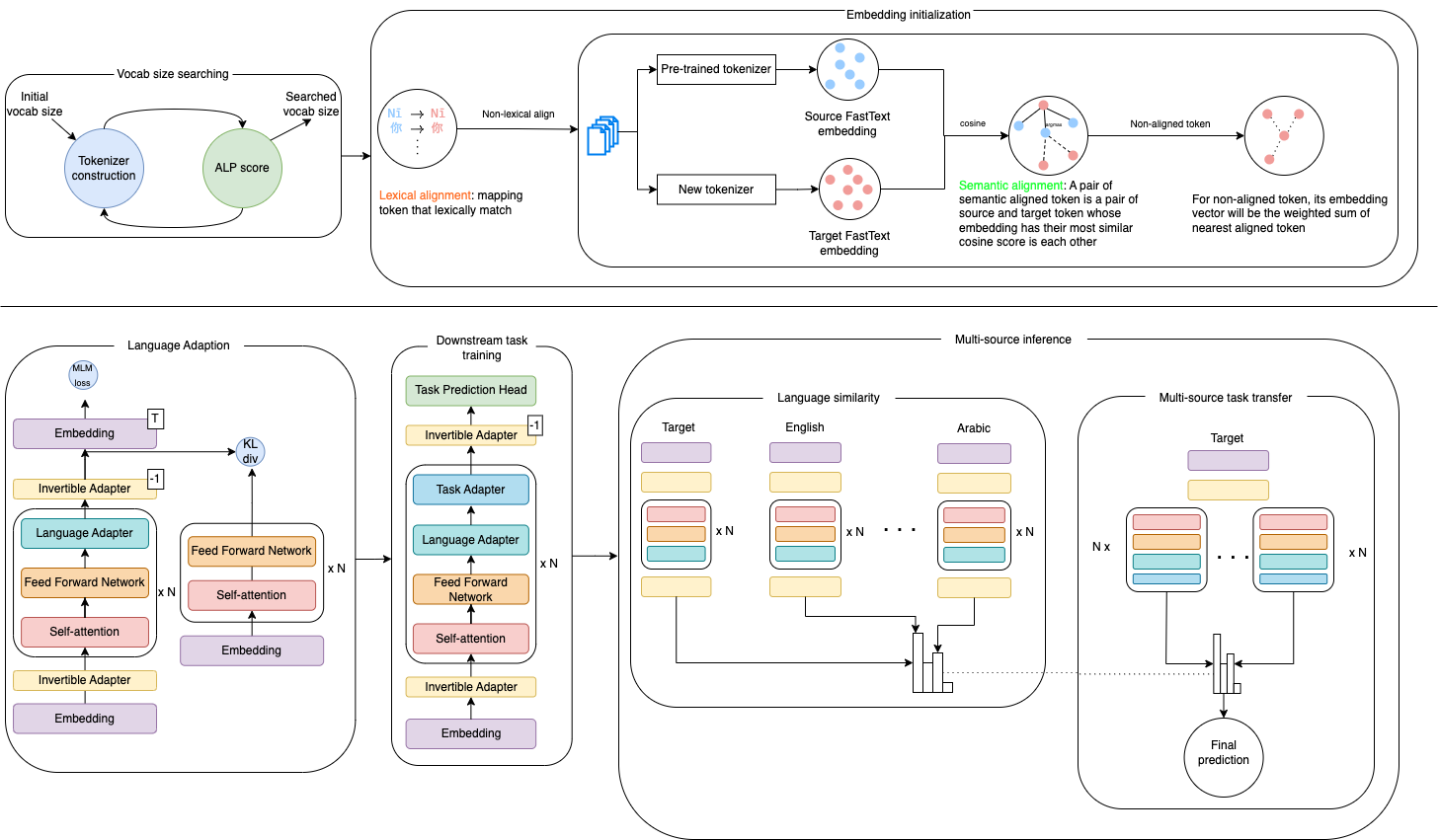}
    \caption{Illustration of UniBridge: UniBridge represents an end-to-end framework for Cross-Lingual Transfer Learning. The framework encompasses various stages, including determining the appropriate vocabulary size, initializing language-specific embedding, adapting the model to new languages, and transferring task knowledge from multiple source languages. This approach aims to harness the power of a multilingual embedding space rather than relying on a single-source transfer language, such as English.}
    \label{fig:pipeline}
\end{figure*}

Our proposed framework includes five stages as illustrated in Figure \ref{fig:pipeline}. In the following section we will detail each stage of the framework: \textbf{1)} Vocabulary size searching, \textbf{2)} Language-specific embedding initialization, \textbf{3)} Model adaptation to new languages not covered in the pre-training data, \textbf{4)} Downstream task training, \textbf{5)} Multi-source transfer downstream task inference.

\subsection{Vocabulary size searching}
Whether training from scratch or starting with a pre-trained language model, every NLP practitioner faces the task of determining the appropriate vocabulary size. Thus, choosing a suitable vocabulary size requires exhaustive searching (i.e., the whole training and testing process is required to determine the best vocabulary size). For UniBridge, the vocabulary is determined by using only CPU and is not time-consuming as it does not require any language model training phases. This is achieved by leveraging the average log probability (ALP, \citet{zheng-etal-2021-allocating}). The algorithm for vocabulary size searching is illustrated by Algorithm \ref{alg:vocab_size_search}.

\begin{algorithm}
\caption{Vocabulary size searching algorithm.}\label{alg:vocab_size_search}
\begin{algorithmic}[1]
\Require $\mathcal{D}$: monolingual data, contains a list of words sentences, $v_i$: initial vocabulary size, $v_m$: maximum vocabulary size that the system should not exceed, $\delta_v$: increased step of vocabulary size, $\epsilon_s$: a threshold for stopping the algorithm.
\State $v \gets v_i$
\State $t \gets $ build tokenizer with vocab size $v$ on $\mathcal{D}$
\State $s_{prev} \gets ALP(\mathcal{D}, t)$
\State $\Delta_s = \infty$
\While{$\Delta_s > \epsilon_s$}
    \State $v \gets v + \delta_v$
    \If{$v > v_m$}
        \State $v \gets v_m$
        \State $t \gets $ build tokenizer with $v$ on $\mathcal{D}$
        \State Break the loop
    \Else
        \State $t \gets $ build tokenizer with $v$ on $\mathcal{D}$
        \State $s_{curr} \gets ALP(\mathcal{D}, t)$
        \State $\Delta_s = s_{curr} - s_{prev}$
        \State $s_{prev} \gets s_{curr}$
    \EndIf
\EndWhile
\State \Return Tokenizer $t$ with vocab size $v$
\end{algorithmic}
\end{algorithm}

The concept of Average Log Probability (ALP) was introduced by \citet{zheng-etal-2021-allocating}, who argue that ALP is related to the effectiveness of subsequent tasks.

\begin{equation}
    ALP(\mathcal{D}, t) = \dfrac{1}{|t(\mathcal{D})|}\sum_{j=1}^{|t(\mathcal{D})|}\sum_{k=1}^{|s_j|}\text{log}p_{uni}(s_j^k)
\end{equation}


For more details, readers are advised to refer to the work of \citet{zheng-etal-2021-allocating}. It is worth noting that although ALP has a high correlation with downstream tasks, the work did not provide a solution to find an optimal vocabulary size. Therefore, in this work, we propose using the `degree of changes' in the ALP score, e.g., $\Delta_s$. Initially, the starting vocabulary size is chosen and the ALP is calculated. Through a series of increases in the vocab size by $\delta_v$, we can calculate the difference between the current ALP and the previous one. Thus, the algorithm will stop when the difference reaches a specific threshold $\epsilon_s$. This threshold indicates that the optimal vocabulary size has been obtained. Continuing to increase the size will result in similar or worse performance. Therefore, we stop the algorithm to maintain the efficiency of training. 
Additionally, our method stands out from traditional grid search by using the `degree of changes' of the ALP score indirectly, rather than directly as in grid search.

\subsection{Language-specific embedding initialization}
\label{sec:lang_embed_init}
When training for a new language, using a randomly initialized embedding can lead to prolonged training times for optimal performance, especially in low-resource settings with a dataset size of around 10K samples. In such cases, strategically initializing the embedding proves to be more effective than a random approach. While FOCUS \citep{dobler-etal-2023-focus} demonstrates the use of a pre-trained LM's embedding for initialization, it depends heavily on a simple lexical overlapped alignment for subsequent stages, thus decreasing the downstream task performance. To address this gap, our approach initializes the new embedding by leveraging the pre-trained LMs in both syntactic and semantic aspects. In the initial stage, we obtain the target tokenizer $t_T$ for the new language, with $t_S$ being the source tokenizer of the pre-trained LMs. Representing the vocabulary sets as $V^T$ and $V^S$ for the target and source tokenizers respectively, and embedding matrices as $E^T[\cdot]$ and $E^S[\cdot]$, we copy the source embedding to the target embedding for the overlapping tokens $O^L = V^T \cap V^S$. This method ensures a seamless integration of knowledge from the pre-trained LMs, addressing both syntactic and semantic aspects of the new language's embedding initialization.

\begin{equation}
    \forall o \in O^L: E^T[o] = E^S[o]
\end{equation}

Although the number of lexical overlapping tokens can be substantial when utilizing the same script, such as Latin or Han, this phenomenon does not extend to unseen scripts. To address this challenge, we define the non-lexical alignment set as $A^L_T = V^T \setminus O^L$ and initiate a search for semantically aligned tokens within this set. Despite languages having different scripts, the underlying meanings often converge on similar definitions. To facilitate this alignment, we train two static embeddings—one for the source tokenizer ($F^S$) and another for the target tokenizer ($F^T$) —using the monolingual dataset $\mathcal{D}$. These embeddings are denoted as $F^S[\cdot]$ for the source tokenizer and $F^T[\cdot]$ for the target tokenizer. For each token $v_i$ in $A^L_T$, we calculate the cosine similarity with every token $v_j$ in $A^L_S = V^S \setminus O^L$, resulting in a matrix $S_{i,j} \in \mathbb{R}^{|A^L_T| \times |A^L_S|}$. A pair of semantically aligned tokens ($v_i$, $v_j$) is defined as a pair of source and target tokens whose embeddings exhibit the highest cosine similarity score to each other, or:

\begin{equation}
    i = \underset{l}{argmax}(S_{l, j})\;\;\text{ and }\;\; j = \underset{l}{argmax}(S_{i, l})
    \label{eq:semantic_align}
\end{equation}

Refer to Equation \ref{eq:semantic_align}, we define $S = \{(i,j) | i = \underset{l}{argmax}(S_{l, j})\text{ and }j = \underset{l}{argmax}(S_{i, l})\}$. Each token that is semantically aligned will have the embedding copied from their counterpart from the source embeddings.

\begin{equation}
    \forall (i, j) \in S: E^T[i] = E^S[j]
\end{equation}

For the remaining non-aligned tokens, $A_T = A^L_T \setminus S_i$ and $A_S = A^L_S \setminus S_j$ where $S_i$, $S_j$ is the set of semantically aligned token of the target and source vocabulary (i.e. $S_i = \{i | (i, j) \in S\}$, $S_j = \{j | (i, j) \in S\}$), we initialize the target embedding using the weighted sum of the aligned target tokens. We compute the cosine similarity between each non-aligned token $a_T \in A_T$ and the set of aligned target tokens (comprising both lexical and semantically aligned tokens) $o_T \in O^L \cup S_i$.

\begin{equation}
    c_{a, o} = \dfrac{F^T[a_T] F^T[o_T]^{\top}}{\|F^T[a_T]\| \cdot \|F^T[o_T]\|}
\end{equation}

To obtain the most similar aligned symbols $o_T$ for a single symbol $a_T$, we use the same approach in \citet{dobler-etal-2023-focus}, using sparsemax \citep{martins-etal-2016-sparsemax} over $c_a$, where $c_a$ is a vector containing all similarity scores from $c_{a,o}$. Sparsemax is a variant softmax, but it assigns zero to low-probability element. By this, we can overcome the problem posted by skew distribution where some tokens has only one or two similar tokens while others have more. The weight $w_{a, o}$ for each aligned token $o_T$ as defined in Equation \ref{eq:init_weighted_sum}.

\begin{equation}
    w_{a, o} = \text{sparsemax}_o(c_a)
    \label{eq:init_weighted_sum}
\end{equation}

We denote $S_a$ as a set of similar aligned tokens, which contains $o_T$ whose probability is non-zero assigned by sparesemax.

\begin{equation}
    S_a = \{o_T \in O^L \cup S_i | w_{a, o} > 0\}
\end{equation}

Using the set \(S_a\) and the weight \(w_{a, o}\), the embedding for the non-aligned token \(a_T\) is calculated as the weighted sum of its most similar aligned tokens.

\vspace{-5mm}
\begin{equation}
    \forall a_T \in A_T: E^T[a_T] = \sum_{o_T \in S_a} w_{a_T, o_T} E^T[o_T]
\end{equation}

\subsection{Model adaptation to new languages \& Downstream task training}
\label{sec:lang_adapt}

Continual pre-training, also known as language adaptation, has proven to be an effective method for enhancing the downstream performance of zero-shot cross-lingual tasks, as demonstrated by studies such as \citet{ke-etal-2023-continual, alabi-etal-2022-adapting, ebrahimi-kann-2021-adapt}. To mitigate the environmental impact and reduce model storage requirements, we opt to pre-train only a portion of the model, aligning with the approach introduced in MAD-X \citep{pfeiffer-etal-2020-mad}.

As in Figure \ref{fig:pipeline}, we made some modifications to the MAD-X configuration. Firstly, we initialize a new embedding for UniBridge which is achieved from previous stages and train the embedding together with adapters while still freezing all the pre-trained LMs' parameters. Secondly, we propose using the KL divergence together with the MLM loss (Appendix \ref{sec:kl_divergence_mlm_loss}). We see that although the frozen parameter in each layer of the pre-trained LMs helps guide the trainable adapters of the new language's embedding representation into the same pre-trained LM's embedding space, MLM is not sufficient as it only enforces the adapter to predict the mask token and this cannot guarantee the new language's representation is the same as multilingual embedding space encoded by the pre-trained LMs. This limitation prohibits the knowledge transferability of task adaptation since task adaptation takes a source language (usually high-resource languages such as English, Chinese, etc) and transfers the task knowledge directly to the target language without any alignment between the two languages. Therefore, we use KL divergence as a regularizer to guide the model not only to learn the language representation well, but also to maintain the same space as the source language in order to achieve better transferability.

\begin{equation}
    \begin{aligned}
        \mathcal{L} &= \mathcal{L}_{MLM}(y, \hat{y}) \\
        &+ \beta D_{KL}(\pi_{UniBridge}(h|x)\|\pi_{PLM}(h|x))    
    \end{aligned}
    \label{eq:main_loss}
\end{equation}

$y$ and $\hat{y}$ are the ground truth and prediction logits of the mask prediction task, respectively. $\pi_{UniBridge}(h|x)$ is the last hidden state of UniBridge, it is the output of the invertible adapter before goes to the linear classification head for masked predicting. $\pi_{PLM}(h|x)$ is the last hidden state of the pre-trained LMs, it is the output of the last Transformer layer, as in Figure \ref{fig:pipeline}, and is the input of the linear classification head for mask predicting.

\subsection{Multi-source transfer downstream task inference}

Instead of using one task adapter from one source language, we propose aggregating the knowledge from multiple source languages to derive a better result. For each target language, we compute the harmony weight or similarity distance between languages. Some libraries such as Lang2Vec \citep{malaviya-etal-2017-learning} provide a similarity score between languages. However it does not cover all the languages. To overcome this problem, we directly use the language model (that UniBridge produced from previous stages) to measure the similarity between languages. In the Appendix \ref{sec:lang2vec-comparison}, we will provide a detailed comparison between Lang2Vec and UniBridge. This analysis will highlight the differences and similarities between the two approaches, offering insights into their respective performances and effectiveness.

For each target language, we collect $K$ samples of parallel sentences from datasets such as Tatoeba \citep{tiedemann-2020-tatoeba} or FLORES-200 \citep{guzman-etal-2019-flores, goyal-etal-2022-flores, nllbteam2022language} between the target language and a set of $N$ source languages.

We denote $\mathcal{D}^T$ as a monolingual dataset extracted from the parallel dataset on the target side, $\mathcal{D}^{S_i}$ is the monolingual dataset extracted from the parallel dataset on the source side of the $i$-th source language. Each sentence is fed into the UniBridge with the corresponding language adapters and obtains a set of hidden states (i.e., output from the invertible adapter).

\begin{equation}
    H_l = \{\pi_{UniBridge}^l(s) | s \in \mathcal{D}^l\}
\end{equation}

$\pi_{UniBridge}^l$ is the UniBridge model which use the $l$ adapter; $\mathcal{D}^l$ is $\mathcal{D}^T$ for the target language and $\mathcal{D}^{S_i}$ for the $i$-th source languages. The inverse $L_2$ distance between the target hidden state $H_t$ for target language $t$ and source hidden state $H_s$ for source language $s$ will be computed.

\begin{equation}
    d_{t, s} = \dfrac{1}{\text{$L_2$-norm}(H_t, H_s)}
\end{equation}

After that, we compute the softmax over the inverse $L_2$ distance to gain the harmony weight between target language $t$ and set of source languages $S = \{s_i\}^N_{i=1}$. 

\begin{equation}
    w_t = \text{softmax}_s(d_{t, s})
\end{equation}

Using this harmony weight, instead of replacing the task adapter for each different source language during inference like MAD-X, we forward through all the task adapters in parallel. The last logit prediction will be the weighted sum of all the logits predicted by each source language weighted by the harmony weight.

\begin{equation}
    \hat{y} = \sum_{s \in S}w_{t, s}\hat{y}_s
\end{equation}

$\hat{y}_s$ is the logit prediction from source language $s$.

The intuition behind the harmony weight is that given a pair of parallel sentence, each sentence is encoded by a different language model. The difference between the hidden states produced by this process turns out to be the difference between languages itself since the sentences convey the same meaning. Therefore, inversing the difference and applying softmax will result in the similarity that we can up-weight for languages, and they could be beneficial to the target language on downstream tasks and, at the same time, down-weight the languages that are distant from the target language. Through our experiment, we show that multi-source inference outperforms single-source transfer and multi-language learning settings.

\section{Experimental setup}
\label{sec:expr_setup}
\textbf{Language set}: The set of source languages are: English, Chinese, Russian, Arabic and Japanese.  For the target languages, we evaluate \textbf{\textit{14}} low-resource languages from WikiANN \citep{rahimi-etal-2019-massively} whose training set consists of only 100 samples per language, \textbf{\textit{9}} low-resource languages from Universal Dependencies (UD) whose training set consists of just few thousands samples per language and \textbf{\textit{10}} languages from the AmericasNLI \citep{ebrahimi-etal-2022-americasnli}.

\textbf{Monolingual data}: For the language adaptation part, we extract from the Wikipedia dataset from HuggingFace \footnote{\url{https://huggingface.co/datasets/graelo/wikipedia}} 10K samples for simulating the low-resource settings, each sample consists of 128 words, for each target language. For source languages, the number of samples is 50K per language to simulate the rich-resource languages. For languages in AmericasNLI, we use one side of the translation dataset from \citet{mager-etal-2021-findings}.

\textbf{Tokenizer}: We use the SentencePiece \citep{kudo-richardson-2018-sentencepiece} to learn the token from the monolingual data with the vocab size determined by our Algorithm \ref{alg:vocab_size_search}.

\textbf{Downstream data}: \textbf{\textit{NER}}: We train UniBridge on the train split of WikiANN for all the source language sets and perform inference for the target language on the test split. \textbf{\textit{POS}}: We train UniBridge from the train split of UD for all the source languages sets. \textbf{\textit{NLI}}: We train UniBridge from the train split of XNLI \citep{conneau-etal-2018-xnli} for English, Chinese, Arabic and Russia due to the missing Japanese set.

\textbf{Baseline}: We evaluate UniBridge against the MAD-X framework and zero-shot cross-lingual fine-tuning using pre-trained language models (LMs). In the zero-shot approach, we fine-tune the entire pre-trained LM on the combined training data of all source languages and then directly infer on the target languages. With MAD-X, we adhere to its standard setup, training on monolingual data. To perform multi-language training, we combine training data from all source languages to train a ``universal'' task adapter. For inference, we swap the language adapter for each target language and integrate the ``universal'' task adapter. For UniBridge, we implement the language adaptation and task training stages as detailed in Section \ref{sec:lang_adapt}. During inference, we combine the task adapters from 5 source languages for multi-source transfer and report the F1 score for \textbf{\textit{NER}} and accuracy score \textbf{\textit{POS}}, \textbf{\textit{NLI}} on the target language's test split.  

The hyperparameters for training, inference as well as the computational resources are given in Appendix \ref{sec:appendix}.

\section{Results and Analysis}

\begin{table*}[t]
    \centering
    \resizebox{1\textwidth}{!}{
    \begin{tabular}{l|cccccccccccccc|c}
          & \textbf{amh} & \textbf{ang} & \textbf{cdo} & \textbf{crh} & \textbf{eml} & \textbf{frr} & \textbf{khm} & \textbf{kan} & \textbf{lij} & \textbf{pbt} & \textbf{san} & \textbf{snd} & \textbf{sin} & \textbf{som} & \textbf{Average}\\
         \hline
         XLM-R & \underline{43.31}& \textbf{52.71}& \underline{22.04}& \underline{44.62}& \underline{40}& \underline{44.17}& 40.69& \underline{45.34}& \underline{40.45}& \underline{46}& \textbf{41.28}& \underline{43.13}& \underline{50.03}& \underline{50.23}& \underline{43.14}\\
         MAD-X (XLM-R) & 39.3& \underline{46.59}& 17.32& 36.63& 33.86& 39.51& \textbf{50}& 45.24& 38.13& 42.66& 19.93& 39.06& 39.55& 49.6& 38.38\\
         UniBridge (XLM-R) & \textbf{49.6}& 43.24& \textbf{42.91}& \textbf{46.03}& \textbf{40.15}& \textbf{50.67}& \underline{42.67}& \textbf{48.72}& \textbf{45.16}& \textbf{46.09}& \underline{29.74}& \textbf{51.32}& \textbf{52.86}& \textbf{54.22}& \textbf{45.95}\\
    \end{tabular}}
    \caption{The results of the F1 Score for every setup with XLM-R as a backbone are showcased in 14 diverse languages of WikiANN. We highlight in \textbf{bold} the highest F1 score and \underline{underline} the second highest of each target language for each backbone model.}
    \label{tab:main_result_xlm}
\end{table*}

\begin{table*}[t]
    \centering
    \resizebox{1\textwidth}{!}{
    \begin{tabular}{l|cccccccccccccc|c}
          & \textbf{amh} & \textbf{ang} & \textbf{cdo} & \textbf{crh} & \textbf{eml} & \textbf{frr} & \textbf{khm} & \textbf{kan} & \textbf{lij} & \textbf{pbt} & \textbf{san} & \textbf{snd} & \textbf{sin} & \textbf{som} & \textbf{Average}\\
         \hline
         mBERT & 12.87& \underline{52.24}& \underline{19.76}& \textbf{47.81}& \textbf{39.71}& \underline{51.3}& 18.46& \underline{42.86}& \underline{45}& 25.86& \textbf{30.71}& 13.61& 2.79& \underline{46.15}& \underline{32.08}\\
         MAD-X (mBERT) & \underline{13.91}& 51.48& 16.22& \underline{46.22}& \underline{39.2}& 45.76& \underline{19.2}& 31.3& 37.35& \underline{29.25}& \underline{22.96}& \underline{20.31}& \underline{12.34}& 37.66& 30.23\\
         UniBridge (mBERT) & \textbf{15.46}& \textbf{53.28}& \textbf{30.42}& 45.67& 36.15& \textbf{54.72}& \textbf{19.49}& \textbf{44.07}& \textbf{45.49}& \textbf{39.33}& 20.55& \textbf{42.36}& \textbf{13.68}& \textbf{62.28} & \textbf{37.35} \\
    \end{tabular}}
    \caption{The results of the F1 Score for every setup with mBERT as a backbone showcased in 14 diverse languages of WikiANN. We highlight in \textbf{bold} the highest F1 score and \underline{underline} the second highest of each target language for each backbone model.}
    \label{tab:main_result_mbert}
\end{table*}

\begin{table*}[t]
    \centering
    \resizebox{0.8\textwidth}{!}{
    \begin{tabular}{l|ccccccccc|c}
          & \textbf{amh} & \textbf{lij} & \textbf{olo} & \textbf{san} & \textbf{snd} & \textbf{sin} & \textbf{tam} & \textbf{tgl} & \textbf{tat} & \textbf{Average}\\
         \hline
         XLM-R & \underline{46.02}& 39.15& 60.69& 32.9& 70.01& \textbf{76.25}& \textbf{85.53}& 67.45& 57.89& 59.54\\
         MAD-X (XLM-R) & \textbf{47.72}& \underline{58.28}& \underline{69.48}& \underline{36.1}& \underline{71.2}& \underline{73.86}& 83.85& \underline{69.01}& \underline{65.83}& \underline{63.88}\\
         UniBridge (XLM-R) & 40.88& \textbf{73.75}& \textbf{81.45}& \textbf{38.94}& \textbf{71.37}& 63.52& 83.5& \textbf{72.62}& \textbf{81.3} & \textbf{67.81}\\
    \end{tabular}}
    \caption{The results of the accuracy for every setup with XLM as a backbone are showcased in 9 diverse languages of UD. We highlight in \textbf{bold} the highest accuracy score and \underline{underline} the second highest of each target language for each backbone model.}
    \label{tab:pos_xlm}
\end{table*}

\begin{table*}[t]
    \centering
    \resizebox{0.8\textwidth}{!}{
    \begin{tabular}{l|ccccccccc|c}
          & \textbf{amh} & \textbf{lij} & \textbf{olo} & \textbf{san} & \textbf{snd} & \textbf{sin} & \textbf{tam} & \textbf{tgl} & \textbf{tat} & \textbf{Average}\\
         \hline
         mBERT & 8.59& \underline{60.66}& \underline{61.49}& 9.35& 20.39& 11.47& \underline{72.93}& \underline{66.3}& \underline{83.2}& 37.82\\
         MAD-X (mBERT) & \underline{13.31}& 50.47& 59.61& \underline{10.88}& \underline{24.93}& \underline{25.68}& 66.61& 55.56& 74.17& \underline{42.36}\\
         UniBridge (mBERT) & \textbf{29.24}& \textbf{65.53}& \textbf{70.65}& \textbf{12.86}& \textbf{66.78}& \textbf{52.61}& \textbf{75.23}& \textbf{70.65}& \textbf{84.16}& \textbf{58.64}\\
    \end{tabular}}
    \caption{The results of the accuracy for every setup with mBERT as a backbone are showcased in 9 diverse languages of UD. We highlight in \textbf{bold} the highest accuracy score and \underline{underline} the second highest of each target language for each backbone model.}
    \label{tab:pos_mbert}
\end{table*}

We present the result of our method and the baselines in Table \ref{tab:main_result_xlm} and \ref{tab:main_result_mbert} for NER task and Table \ref{tab:pos_xlm} and \ref{tab:pos_mbert} for POS tagging task. We report the NLI results in Table \ref{tab:nli_xlm} and \ref{tab:nli_mbert} in Appendix \ref{sec:nli}. UniBridge outperforms strong baselines such as whole model fine-tuned (XLM-R, mBERT) and MAD-X framework by a large margin, i.e, for the XLM-R model, we outperform 11 over 14 languages. For POS tagging task, we outperform both baselines with two different backbone models. We also see this trend in NLI task (Appendix \ref{sec:nli}). This highlights the effect of leveraging multiple source languages during inference to help make better decisions since each language contributes knowledge that benefits the model at prediction. Meanwhile, multi-training offers a more robust performance but also introducing more difficulties during training. The fact that UniBridge outperforms strong baselines such as whole fine-tuned model indicates that given a small monolingual and lightweight adaptation using adapters, we can significantly improve the cross-lingual tasks for uncovered languages. Compared to MAD-X, our approach differs from the use of a new embedding layer. For unseen languages, using a more specific layer of embedding can remarkably enhance the performance. Even though MAD-X already uses the invertible adapters as a component to adapt embedding layer to unseen languages, these components may not sufficient for rare languages with unseen scripts such as Amharic (\textbf{amh}), Khmer (\textbf{khm}), Kanada (\textbf{kan}). In addition, to evaluate UniBridge with large (decoder-style) Language Models (LLMs), we expanded our experiments beyond XLM-R and mBERT to include mGPT \citep{shliazhko-etal-2024-mgpt} and mBART \citep{liu-etal-2020-multilingual-denoising}. This extension provides a more robust assessment of UniBridge's effectiveness across different model types, demonstrating its versatility and potential in leveraging various LLM architectures for improved language representation. The results are presented in Appendix \ref{sec:mgpt-mbart-llms}, showcasing the comparative performance and strengths of UniBridge in diverse settings.

Although UniBridge can successfully improve cross-lingual generalization, there are still some inconsistencies in the performance of a language on different tasks, e.g., Amharic (\textbf{amh}), Ligurian (\textbf{lij}), and Sanskrit (\textbf{san}) on NER and POS tasks. We hypothesize that the inconsistency arises from the misalignment in the subspace between the language adapter and the task adapter. One approach to mitigate this misalignment is to regularize the representation so that the newly learned representation is shared between the source and target languages. UniBridge leverages KL divergence as a regularization approach. This may not be sufficient to completely resolve the inconsistency, but given our constrained resources, KL divergence fits our requirements well. We leave other advanced methods, such as optimal transport or contrastive learning, for future work.

\section{Ablation study}

\subsection{Contribution of each component}

We study the contribution of each UniBridge component independently to investigate the critical components of each module. To remove KL divergence, we simply remove the KL loss from equation \ref{eq:main_loss}, keeping only MLM loss. To remove the embedding initialization component, we randomly initialize the embedding drawn from the Xavier normal distribution \cite{pmlr-v9-glorot10a}. To remove the vocab size search component, we fix the vocab size to 10k for every target language and use SentencePiece \citep{kudo-richardson-2018-sentencepiece}. To remove the multi-source transfer, we consider English as the single source language transferred due to its wide use in many cross-lingual transfer works.

\begin{table}[t]
    \centering
    \resizebox{0.46\textwidth}{!}{
    \begin{tabular}{l|cc}
        & \textbf{XLM-R} & \textbf{mBERT} \\
        \hline
        UniBridge & 45.95$\pm$6.28 & 37.35$\pm$15.38 \\
        \hdashline
        - KL Divergence & 46.87$\pm$7.02 &  34.78$\pm$17.48\\
        - Embedding initialization & 6.56$\pm$6.11 & 10.21$\pm$8.72 \\
        - Vocab size searching & 45.48$\pm$7.54 & 30.59$\pm$14.55 \\
        - Multi-source transfer & 42.05$\pm$9.91 & 25.66$\pm$12.3 \\
    \end{tabular}}
    \caption{The performance of UniBridge when removing each component independently. Here, each removed component are indicating by the minus sign (-). For each removed components, other components are remained the same as the default configuration.}
    \label{tab:ablation_study}
\end{table}

We report the mean and standard deviation of the F1 scores between 14 languages of 2 backbone models when applying UniBridge and the components removed from UniBridge in Table \ref{tab:ablation_study}, details of each language can be found in Appendix \ref{sec:detail_ablation}. Among components, \textbf{embedding initialization} plays the most critical role since removing it, we experience performance drops of about 39 and 20 for XLM-R and mBERT, respectively. For \textbf{multi-source transfer} component, mBERT experiences a larger drop with 11 F1 drop while XLM-R is down from 45 to 42. However, the standard deviation when removing multi-source transfer is larger than that of UniBridge (XLM-R), indicating that multi-source benefits more languages compared to single language transferred. Although removing \textbf{KL divergence} off the XLM-R improves its performance by 1 F1 score, the standard deviation increases by 1 score. Thus, KL divergence benefits languages in maintaining a more stable improvement among languages. On the other hand, removing KL divergence while using mBERT as a backbone model hurts the performance and drops 3 F1 scores. In order to clarify the effectiveness of KL-Divergence in the other model, we conducted experiments in Appendix \ref{sec:kl-divergence}. \textbf{Vocab size searching} with dynamic vocab size significantly improves the performance for mBERT backbone with an improvement of the 7 F1 score. This implies that different languages should be applied differently and dynamically to best adapt to their linguistic features. 

\subsection{Vocabulary size}

In this section, we contrast our approach with a novel technique for vocabulary initialization called EXTEND \citep{wang-etal-2020-extending}. EXTEND operates by initially expanding mBERT's vocabulary to accommodate the new language and then proceeding with pre-training on this language. In our comparison, EXTEND undergoes full fine-tuning for the MLM pre-training task. Subsequently, EXTEND is further fine-tuned using the monolingual data of each target language. Despite its extensive fine-tuning and high computational requirements, EXTEND does not perform satisfactorily on NER in comparison to UniBridge, as illustrated in Table \ref{tab:mbert_extend_bert} in Appendix \ref{sec:unibridge_vs_extend}. UniBridge offers a much lighter and faster alternative, employing adapters for cross-lingual transfer learning. The lightweight and rapid nature of UniBridge significantly enhances the effectiveness of our method. Furthermore, we present an elaborate Table \ref{tab:ablation_study_3} containing various vocabulary sizes for each target language in the Appendix \ref{sec:vocab_search_res}. Regarding the lexical similarity of subwords in the vocabulary, we offer illustrations of subwords that exhibit similarity in both mBERT and XLM-R. These examples can be found in Figure \ref{fig:lexical_similarity} within Appendix \ref{sec:similar_lexical_token}.

\subsection{ALP Threshold}

\begin{figure}[ht]
    \centering
    \includegraphics[width=\linewidth]{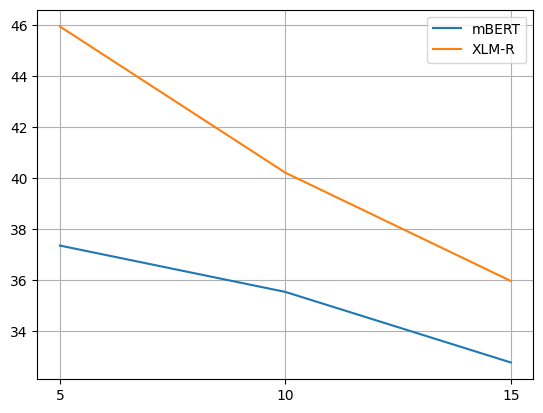}
    \caption{Mean F1-Score across various ALP thresholds.}
    \label{fig:alp_threshold_mean_f1_scores}
\end{figure}

We conducted experiments using different ALP thresholds to identify the most effective one. We tested threshold values such as 5.0, 10.0, and 15.0 during the pre-training process of UniBridge. In essence, raising the threshold led to a decrease in vocabulary size as the algorithm ended prematurely. As a result, we noticed a decrease in the F1-Score of mBERT and XLM-R as the threshold values increased, as illustrated in Figure \ref{fig:alp_threshold_mean_f1_scores}.

More experiments and ablation study can be found in Appendix \ref{sec:more_experiments}.

\section{Related works}
\label{sec:related-works}

\textbf{Dynamic vocabulary size}: It is common among NLP practitioners that the vocabulary size is considered a hyperparameter and requires manual settings. Algorithms such as BPE \citep{gage1994new}, WordPiece \citep{wu2016googles}, SentencePiece \citep{kudo-richardson-2018-sentencepiece} focus more on how to build a set of predefined number of tokens that statistically retrieved from the monolingual dataset. Some work such as VoCAP \citep{zheng-etal-2021-allocating}, XLM-V \citep{liang-etal-2023-xlm} proposed algorithm to dynamically assign a vocabulary size for each language during the multilingual training. However, in monolingual training, there are some few works concerning this problem such as BPE-dropout \citep{provilkov-etal-2020-bpe}, VOLT \citep{xu-etal-2021-vocabulary} learns to have an optimal vocab size via reducing the original vocab size using optimal transport as in VOLT or randomly removing merge in BPE-dropout. 

\textbf{Initialization}: \citet{artetxe-etal-2020-cross} proposed to randomly initialize the new embedding for new language adaptation. Meanwhile, \citet{wang-etal-2020-extending}, \citet{chau-etal-2020-parsing}, \citet{pfeiffer-etal-2021-unks} leverage the lexical similarity between the old vocabulary and the new vocabulary to initialize the embedding. On the other hand, there are works that explore the semantic space for initialization. SMALA \citep{vernikos-popescu-belis-2021-subword-mapping} directly finds the aligned token through the highest cosine similarity score. WECHSEL \citep{minixhofer-etal-2022-wechsel} and FOCUS \citep{dobler-etal-2023-focus} use static embedding to find aligned tokens.

\textbf{Multi-source transfer}: Single-source transfer, especially, English-as-the-source-language receives many attentions. \citet{artetxe-etal-2020-cross}, \citet{ansell-etal-2021-mad-g}, \citet{tu-etal-2022-prompt} leverages the multilingual backbone model, fine-tune on English downstream task and perform zero-shot transfer on target language test's set. Until recently, researchers have pointed out that using a multilingual training set is more beneficial compared to a single language. DeMuX \citep{khanuja2023demux} investigates the dataset level to accumulate examples that best benefit transferring using active learning. \citet{dossou-etal-2022-afrolm}, \citet{ogunremi-etal-2023-mini} pre-train on the multilingual African dataset before distilling knowledge to downstream tasks.

\section{Conclusion}

In this paper, we investigate Cross-Lingual Transfer Learning, focusing on languages with constrained resources. Our contribution lies in an algorithm that autonomously determines the optimal vocabulary size for a new language, informed by its monolingual corpus, and an innovative method for initializing a new embedding matrix, drawing from both semantic and lexical facets of the pre-trained language models. Additionally, we introduce a novel technique for aggregating multi-source transfer learning, enhancing the efficacy of cross-lingual transfer tasks. Our empirical tests demonstrate the adaptability of our method across different models, yielding significant enhancements in performance. A thorough investigation of key elements highlights UniBridge's effectiveness in various situations, offering an in-depth understanding of the robustness of our approach.

\section*{Limitation}

UniBridge is trained on the extracted Wikipedia with some heuristic noise filtering. However, we believe that further pre-processing such as language identification and noise filtering pipeline could further produce higher-quality monolingual data, which potentially improve the language adaptation stage. UniBridge incorporates the use of adapter to perform cross-lingual generalization, while this leverages the modular characteristic of adapter, it also inherited some limitation of the adapter itself \citep{kunz-holmstrom-2024-impact, alabi2024hidden}.

\bibliography{custom}

\appendix

\section{Why KL Divergence and MLM Loss work?}
\label{sec:kl_divergence_mlm_loss}

For the KL-Divergence effect, in contrast to MAD-X, UniBridge incorporates a novel training embedding. This results in the pre-task adapter representation being more inclined to reflect the characteristics of the target language compared to solely using the language adapter in MAD-X. Consequently, this introduces a misalignment between the language adapter and the task adapter, as each represents a different language.

In our research, we opt for KL-Divergence to regulate the representation to ensure it is shared across both the source and target languages \cite{kim2021comparing}. KL-Divergence requires less computational resources compared to other methods like layer-wise regularization or optimal transport.

For the MLM Loss effect, it is extremely effective in training encoder-only LMs because it encourages the model to learn rich contextual representations of language and facilitates effective pre-training.

In MLM, a portion of the input tokens is randomly masked, and the model is trained to predict these masked tokens based on the context provided by the surrounding tokens. This forces the model to learn contextual representations of words and phrases of that target language \cite{sinha-etal-2021-masked}. Moreover, by randomly masking tokens, MLM introduces noise into the training process, which can prevent overfitting and encourage the model to learn more generalizable features of the data.

\section{Similar tokens between pre-trained LM and UniBridge specific tokenizer}
\label{sec:similar_lexical_token}

\begin{figure}[ht]
    \centering
    \includegraphics[width=\linewidth]{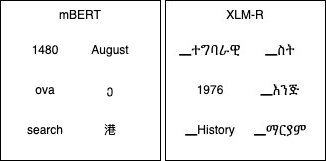}
    \caption{Illustrations of subwords exhibiting similarity in both mBERT and XLM-R.}
    \label{fig:lexical_similarity}
\end{figure}

We illustrate the similar tokens between pre-trained LM and UniBridge specific tokenizer in Figure \ref{fig:lexical_similarity}.

\section{Computational resources and hyperparameter for training, inference}
\label{sec:appendix}

All experiments are conducted on T4 machines. Training the UniBridge's language adapter takes approximately 2.5 hours on a single T4 machine with a batch size of 16. Separately, training UniBridge's task adapter, takes about 0.5 hours per source language on a single T4 machine with a batch size of 16.

We present the hyperparameters for training and inference for UniBridge and all the baselines' configurations in Table \ref{tab:hyper_vocab_size}, \ref{tab:hyper_embed_init}, \ref{tab:hyper_lang_train}, \ref{tab:hyper_task_train}, \ref{tab:hyper_zero_shot_plm}, and \ref{tab:hyper_infer}.

\begin{table}
    \centering
    \resizebox{0.45\textwidth}{!}{
    \begin{tabular}{c|c}
        \textbf{Hyperparameter} & \textbf{Value} \\
        \hline
        Initial vocab size $v_i$ & 7000 \\
        Maximum vocab size $v_m$ & 60000 \\
        Increased step of vocab size $\delta_v$ & 1000 \\
        Threshold for stopping the algorithm $\epsilon_v$ & 5.0 \\
    \end{tabular}}
    \caption{The hyperparameter for vocabulary size searching process.}
    \label{tab:hyper_vocab_size}
\end{table}

\begin{table}
    \centering
    \begin{tabular}{c|c}
        \textbf{Hyperparameter} & \textbf{Value} \\
        \hline
        Static embedding model & FastText \\
        Static embedding dimension & 300 \\
        Number of epoch of training & 3 \\
    \end{tabular}
    \caption{The hyperparameter for embedding initialization stage.}
    \label{tab:hyper_embed_init}
\end{table}

\begin{table}
    \centering
    \resizebox{0.5\textwidth}{!}{
    \begin{tabular}{c|c}
        \textbf{Hyperparameter} & \textbf{Value} \\
        \hline
        Invertible adapter reduced factor & 2 \\
        Language adapter reduced factor & 2 \\
        KL divergence weight $\beta$ & 1.0 \\
        Masked probability & 0.15 \\
        Number of epochs trained & 50 \\
        Batch size & 32 \\
        Learning rate & \{5e-5, 2e-4, 5e-4, 1e-3\} \\
    \end{tabular}}
    \caption{The hyperparameter for language adaptation training. The adapter dimension is dynamically determined by reducing the Transformer's hidden size by a factor of reduced factor. Each language has a different proportion in the pre-trained LMs' knowledge; therefore, to have an optimal language adaptation, different learning rate for different language is required.}
    \label{tab:hyper_lang_train}
\end{table}

\begin{table}
    \centering
    \begin{tabular}{c|c}
        \textbf{Hyperparameter} & \textbf{Value} \\
        \hline
        Task adapter reduced factor & 16 \\
        Number of epochs trained & 11 \\
        Batch size & 32 \\
        Learning rate & \{5e-4, 1e-3\}
    \end{tabular}
    \caption{The hyperparameter for task adaptation. Each language has a different proportion in the pre-trained LMs' knowledge; therefore, to have an optimal language adaptation, a different learning rate for different languages is required.}
\label{tab:hyper_task_train}.
\end{table}

\begin{table}
    \centering
    \begin{tabular}{c|c}
        \textbf{Hyperparameter} & \textbf{Value} \\
        \hline
        Number of epochs trained & 10 \\
        Batch size & 32 \\
        Learning rate & 1e-5 \\
    \end{tabular}
    \caption{The configuration of the pre-trained LMs' fine-tuning on source downstream task and zero-shot transfer to target language.}
    \label{tab:hyper_zero_shot_plm}
\end{table}

\begin{table}
    \centering
    \begin{tabular}{c|c}
        \textbf{Hyperparameter} & \textbf{Value} \\
        \hline
        Number of parallel sentences $K$ & 10 \\
    \end{tabular}
    \caption{The configuration for multi-source inference.}
    \label{tab:hyper_infer}
\end{table}

\section{More experiments and ablation study}
\label{sec:more_experiments}

\subsection{Performance of UniBridge on NLI task}
\label{sec:nli}

We report the performance of UniBridge on the AmericasNLI dataset in Tables \ref{tab:nli_xlm} and \ref{tab:nli_mbert}.

\begin{table*}[t]
    \centering
    \resizebox{0.9\textwidth}{!}{
    \begin{tabular}{l|cccccccccc|c}
          &  \textbf{aym} &  \textbf{bzd} &  \textbf{cni} &  \textbf{grn} &  \textbf{hch} &  \textbf{nah} &  \textbf{oto} &  \textbf{quy} &  \textbf{shp} &  \textbf{tar}  & \textbf{Average}\\
         \hline
         XLM-R & 36.26 & \textbf{38.53} & 36.4 & 37.33 & \textbf{37.33} & 39.43 & \underline{36.89} & \underline{37.6} & \underline{35.86} & 34.66 & 37.03 \\
         MAD-X (XLM-R) & \underline{39.46} & \underline{36.8} & \underline{38.93} & \underline{39.73} & \underline{35.86} & \underline{40.78} & 33.42 & 37.46 & \textbf{39.06} & \textbf{36.53} & \underline{37.80} \\
         UniBridge (XLM-R) & \textbf{52.13} & \underline{36.8} & \textbf{40.26} & \textbf{59.59} & \underline{35.86} & \textbf{46.88} & \textbf{42.38} & \textbf{59.86} & 35.6 & \underline{36.4} & \textbf{44.58} \\
    \end{tabular}}
    \caption{The results of the accuracy score for every setup with XLM as a backbone showcased in 10 diverse languages of AmericasNLI. We highlight in \textbf{bold} the highest accuracy and \underline{underline} the second highest of each target language for each backbone model.}
    \label{tab:nli_xlm}
\end{table*}

\begin{table*}[t]
    \centering
    \resizebox{0.9\textwidth}{!}{
    \begin{tabular}{l|cccccccccc|c}
          &  \textbf{aym} &  \textbf{bzd} &  \textbf{cni} &  \textbf{grn} &  \textbf{hch} &  \textbf{nah} &  \textbf{oto} &  \textbf{quy} &  \textbf{shp} &  \textbf{tar}  & \textbf{Average}\\
         \hline
         mBERT & \underline{33.33} & \textbf{33.33} & \underline{33.33} & 33.33 & 33.33 & \underline{33.33} & \underline{33.2} & 33.28 & \underline{33.33} & \underline{33.33} & 33.31 \\
         MAD-X (mBERT) & 33.06 & \textbf{33.33} & \textbf{34.4} & \underline{33.46} & \underline{34} & \underline{33.33} & \textbf{33.42} & \underline{33.73} & 32.93 & 33.2 & \underline{33.54} \\
         UniBridge (mBERT) & \textbf{35.73} & \textbf{33.33} & \underline{33.33} & \textbf{37.46} & \textbf{34.4} & \textbf{36.31} & \textbf{33.42} & \textbf{36.66} & \textbf{34.4} & \textbf{34.4} & \textbf{34.94} \\
    \end{tabular}}
    \caption{The results of the accuracy score for every setup with mBERT as a backbone showcased in 10 diverse languages of AmericasNLI. We highlight in \textbf{bold} the highest accuracy and \underline{underline} the second highest of each target language for each backbone model.}
    \label{tab:nli_mbert}
\end{table*}

\subsection{UniBridge v.s. Lang2Vec}
\label{sec:lang2vec-comparison}

Our method's reliance on parallel data enables it to capture typological similarities as well as syntactic and semantic relationships between languages. By utilizing parallel sentences, we can develop more nuanced representations that reflect the intricacies of language structures and meanings.

Moreover, the quality and coverage of typological databases can be inconsistent. Although these databases are available for many languages, they often lack completeness and accuracy. In contrast, parallel corpora, while more challenging to obtain, provide direct evidence of language similarities and differences in real-world contexts. Additionally, our method has shown superior performance compared to Lang2Vec in the experiments conducted on the WikiANN dataset in Table \ref{tab:lang2vec-unibridge-comparison-1} and \ref{tab:lang2vec-unibridge-comparison-2}.

\begin{table*}[t]
    \centering
    \resizebox{0.9\textwidth}{!}{
    \begin{tabular}{l|cccccccccccccc|c}
          & \textbf{amh} & \textbf{ang} & \textbf{cdo} & \textbf{crh} & \textbf{eml} & \textbf{frr} & \textbf{khm} & \textbf{kan} & \textbf{lij} & \textbf{pbt} & \textbf{san} & \textbf{snd} & \textbf{sin} & \textbf{som} & \textbf{Average} \\
         \hline
         lang2vec (XLM-R) & 30.19 & \textbf{45.51} & 36.28 & 45.8 & 32.23 & 41.72 & 37.75 & 47.45 & 31.67 & 40.05$^\dagger$ & \textbf{49.79} & 44.84 & 48.95 & 42.17 & 38.03 \\
         UniBridge (XLM-R) & \textbf{49.6} & 43.24 & \textbf{42.91} & \textbf{46.03} & \textbf{40.15} & \textbf{50.67} & \textbf{42.67} & \textbf{48.72} & \textbf{45.16} & \textbf{46.09} & 29.74 & \textbf{51.32} & \textbf{52.86} & \textbf{54.22} & \textbf{45.96} \\
    \end{tabular}}
    \caption{Comparison between Lang2Vec and UniBridge using the XLM-R backbone on the WikiANN dataset. The highest F1 score for each target language is highlighted in \textbf{bold}. The average value for each row is calculated in the last column. $^\dagger$: The language Pashto (\textbf{pbt}) does not exist in the dictionary of lang2vec thus we set the average weight for it, e.g. 0.2 for every source language.}
    \label{tab:lang2vec-unibridge-comparison-1}
\end{table*}

\begin{table*}[t]
    \centering
    \resizebox{0.9\textwidth}{!}{
    \begin{tabular}{l|cccccccccccccc|c}
          & \textbf{amh} & \textbf{ang} & \textbf{cdo} & \textbf{crh} & \textbf{eml} & \textbf{frr} & \textbf{khm} & \textbf{kan} & \textbf{lij} & \textbf{pbt} & \textbf{san} & \textbf{snd} & \textbf{sin} & \textbf{som} & \textbf{Average} \\
         \hline
         lang2vec (mBERT) & 8.76 & 26.85 & \textbf{32.25} & 34.0 & 21.23 & 16.0 & \textbf{26.85} & 37.38 & 27.51 & 28.32$^\dagger$ & 12.12 & 11.34 & 12.57 & 35.42 & 23.61 \\
         UniBridge (mBERT) & \textbf{15.46} & \textbf{53.28} & 30.42 & \textbf{45.67} & \textbf{36.15} & \textbf{54.72} & 19.49 & \textbf{44.07} & \textbf{45.49} & \textbf{39.33} & \textbf{20.55} & \textbf{42.36} & \textbf{13.68} & \textbf{62.28} & \textbf{37.35} \\
    \end{tabular}}
    \caption{Comparison between Lang2Vec and UniBridge using the mBERT backbone on the WikiANN dataset. The highest F1 score for each target language is highlighted in \textbf{bold}. The average value for each row is calculated in the last column. $^\dagger$: The language Pashto (\textbf{pbt}) does not exist in the dictionary of lang2vec thus we set the average weight for it, e.g. 0.2 for every source language.}
    \label{tab:lang2vec-unibridge-comparison-2}
\end{table*}

\subsection{Detail performance of each factor}
\label{sec:detail_ablation}

We present the detailed performance of UniBridge on 14 languages on NER task when removing the contributed components in Table \ref{tab:ablation_study_1} and \ref{tab:ablation_study_2}.

\begin{table*}
    \centering
    \resizebox{0.9\textwidth}{!}{
    \begin{tabular}{l|cccccccccccccc}
          & \textbf{am} & \textbf{ang} & \textbf{cdo} & \textbf{crh} & \textbf{eml} & \textbf{frr} & \textbf{km} & \textbf{kn} & \textbf{lij} & \textbf{ps} & \textbf{sa} & \textbf{sd} & \textbf{si} & \textbf{so} \\
         \hline
         \hline
         UniBridge (XLM-R) & \textbf{49.6}& 43.24& 42.91& 46.03& 40.15& 50.67& \textbf{42.67}& 48.72& \textbf{45.16}& 46.09& 29.74& \textbf{51.32}& \textbf{52.86}& 54.22\\
         \hdashline
         - KL Divergence & 47.66& 45.61& 47.1& 45.91& 37.78& \textbf{58.1}& 40& \textbf{50}& 43.92& \textbf{49.61}& \textbf{31.91}& 50.74& 51.1& \textbf{56.79}\\
         - Embedding initialization & 6.64& 1.23& 0.59& 2.43& 1.56& 2.49& 15.53& 11.32& 12.32& 2.32& 1.15& 15.38& 2.87& 15.95\\ 
         - Vocab size searching & 36.13& \textbf{57.14}& \textbf{47.37}& \textbf{47.54}& \textbf{42.91}& 54.95& 39.65& 45.76& 42.75& 46.44& 28.06& 47.35& 47.51& 53.11\\ 
         - Multi-source transfer & 40.58& 56.13& 36.68& 45.49& 35.96& 57.14& 32.67& 45.53& 39.23& 33.77& 22.93& 39.27& 47.37& 55.97\\
    \end{tabular}}
    \caption{The detailed performance of UniBridge based on backbone model XLM-R when removing contributed components.}
    \label{tab:ablation_study_1}
\end{table*}

\begin{table*}
    \centering
    \resizebox{0.9\textwidth}{!}{
    \begin{tabular}{l|cccccccccccccc}
          & \textbf{am} & \textbf{ang} & \textbf{cdo} & \textbf{crh} & \textbf{eml} & \textbf{frr} & \textbf{km} & \textbf{kn} & \textbf{lij} & \textbf{ps} & \textbf{sa} & \textbf{sd} & \textbf{si} & \textbf{so} \\
         \hline
         \hline
         UniBridge (mBERT) & 15.46& \textbf{53.28}& \textbf{30.42}& 45.67& \textbf{36.15}& 54.72& \textbf{19.49}& \textbf{44.07}& 45.49& 39.33& 20.55& \textbf{42.36}& 13.68& \textbf{62.28}\\
         \hdashline
         - KL Divergence & 2.42& 52.07& 25.52& 42.97& 32.7& \textbf{55.56}& 19.29& 40.69& \textbf{46.15}& \textbf{40.15}& 16.43& 40.14& 11.58& 61.26\\
         - Embedding initialization & 6.58& 3.59& 23.53& 12.35& 9.84& 27.75& 2.34& 11.06& 13.04& 7.61& 1.54& 1.1& 1.2& 21.36\\ 
         - Vocab size searching & 0.15 & 43.82& 17.78& \textbf{48.8}& 32.74& 47.58& 16.74& 33.61& 34.92& 29.06& \textbf{23.32}& 35.99& \textbf{15.18}& 48.51\\ 
         - Multi-source transfer & \textbf{25.08}& 47.21& 15.68& 30.72& 19.86& 41.95& 9.33& 29.37& 29.86& 21.73& 11.26& 22.84& 10.69& 43.66\\ 
    \end{tabular}}
    \caption{The detailed performance of UniBridge based on backbone model mBERT when removing contributed components.}
    \label{tab:ablation_study_2}
\end{table*}

\subsection{Effectiveness of KL Divergence}
\label{sec:kl-divergence}

In contrast to MAD-X, UniBridge employs a novel training embedding, leading to a pretask adapter representation that better captures the characteristics of the target language than solely using the language adapter in MAD-X. To ensure that the output representation is shared between both source and target languages, we use KL-Divergence. This approach is less computationally intensive than methods such as layer-wise regularization or optimal transport (Section \ref{sec:related-works}).

To assess the effectiveness of using KL-Divergence within UniBridge, we conducted extensive tests on an alternative Language Model, such as mBART, using the WikiANN dataset. The results in Table \ref{tab:effective-kl-divergence} indicate that KL-Divergence significantly contributes to the overall performance of UniBridge, enhancing its effectiveness considerably.

\begin{table*}[t]
    \centering
    \resizebox{0.9\textwidth}{!}{
    \begin{tabular}{l|cccccccccccccc|c}
          & \textbf{amh} & \textbf{ang} & \textbf{cdo} & \textbf{crh} & \textbf{eml} & \textbf{frr} & \textbf{khm} & \textbf{kan} & \textbf{lij} & \textbf{pbt} & \textbf{san} & \textbf{snd} & \textbf{sin} & \textbf{som} & \textbf{Average}\\
         \hline
         mBART & 19.19& 15.47& 10.46& 9.1& 14.92& 18.86& 13.16& 15.52& 6.22& 11.45& 19.31& 16.68& 13.21& 14.04& 14.11\\
         MAD-X (mBART) & \underline{67.03}& 51.24& 56.57& 29.73& 39.13& 51.5& 28.79& \underline{43.52}& \textbf{49.72}& \underline{45.25}& 51.85& \underline{58.64}& \textbf{60.33}& 51.2& 48.89\\
         UniBridge without KL-Divergence (mBART) & 53.76& \underline{60.7}& \underline{62.4}& \textbf{65.67}& \underline{66.27}& \underline{56.08}& \underline{33.43}& 39.13& 42.67& 33.13& \underline{59.52}& 45.91& 58.9& \underline{52.02}& \underline{52.11}\\
         UniBridge (mBART) & \textbf{69.15}& \textbf{67.5}& \textbf{67.89}& \underline{61.91}& \textbf{67.14}& \textbf{57.07}& \textbf{41.74}& \textbf{48.37}& \underline{44.1}& \textbf{52.47}& \textbf{60.99}& \textbf{59.12}& \underline{59.29}& \textbf{54.48}& \textbf{57.94}\\
    \end{tabular}}
    \caption{Various configurations with the mBART backbone on the WikiANN dataset. We highlight in \textbf{bold} the highest F1 score and \underline{underline} the second highest of each target language for each backbone model.}
    \label{tab:effective-kl-divergence}
\end{table*}

\subsection{UniBridge v.s. EXTEND}
\label{sec:unibridge_vs_extend}

We report the result on NER task compared between UniBridge and EXTEND method in Table \ref{tab:mbert_extend_bert}.

\begin{table*}[t]
    \centering
    \resizebox{0.9\textwidth}{!}{
    \begin{tabular}{l|cccccccccccccc|c}
          & \textbf{amh} & \textbf{ang} & \textbf{cdo} & \textbf{crh} & \textbf{eml} & \textbf{frr} & \textbf{khm} & \textbf{kan} & \textbf{lij} & \textbf{pbt} & \textbf{san} & \textbf{snd} & \textbf{sin} & \textbf{som} & \textbf{Average}\\
         \hline
         mBERT & \underline{12.87}& 52.24& 19.76& \textbf{47.81}& \textbf{39.71}& \underline{51.3}& 18.46& \underline{42.86}& 45& \underline{25.86}& \textbf{30.71}& 13.61& 2.79& 46.15& 32.08\\
         EXTEND (mBERT) & 10.25& \textbf{60.66}& \underline{26.95}& 42.58& 30.42& 29.71& \textbf{22.04}& 35.41& \textbf{48.63}& 21.16& 14.27& \textbf{49.94}& \underline{11.45}& \underline{50.78}& \underline{32.45}\\
         UniBridge (mBERT) & \textbf{15.46}& \underline{53.28}& \textbf{30.42}& \underline{45.67}& \underline{36.15}& \textbf{54.72}& \underline{19.49}& \textbf{44.07}& \underline{45.49}& \textbf{39.33}& \underline{20.55}& \underline{42.36}& \textbf{13.68}& \textbf{62.28} & \textbf{37.35} \\
    \end{tabular}}
    \caption{The results of the F1 Score for every setup with mBERT as a backbone showcased in 14 diverse languages of WikiANN. We highlight in \textbf{bold} the highest F1 score and \underline{underline} the second highest of each target language for each backbone model.}
    \label{tab:mbert_extend_bert}
\end{table*}

\subsection{Vocabulary searching result of UniBridge}
\label{sec:vocab_search_res}

We report the searched size of the Algorithm \ref{alg:vocab_size_search} for each language in Table \ref{tab:ablation_study_3}.

\begin{table*}
    \centering
    \resizebox{0.9\textwidth}{!}{
    \begin{tabular}{l|cccccccccccccc}
          & \textbf{am} & \textbf{ang} & \textbf{cdo} & \textbf{crh} & \textbf{eml} & \textbf{frr} & \textbf{khm} & \textbf{kan} & \textbf{lij} & \textbf{pbt} & \textbf{san} & \textbf{snd} & \textbf{sin} & \textbf{som} \\
         \hline
         \hline
         UniBridge & 19k& 19k& 10k& 8k& 8k& 18k& 51k& 27k& 20k& 16k& 31k& 14k& 20k& 26k\\ 
    \end{tabular}}
    \caption{The approximate vocabulary sizes of each target language.}
    \label{tab:ablation_study_3}
\end{table*}

\subsection{UniBridge v.s. FOCUS}
\label{sec:unibridge_focus}

We compared UniBridge initialization and FOCUS initialization. For UniBridge, the whole pipeline is kept the same as discussed in the paper. For FOCUS \citep{dobler-etal-2023-focus}, we replace step 2 discussed in Section \ref{sec:lang_embed_init} with FOCUS initialization pipeline while other steps are kept the same as UniBridge. We report the results on NER task in Table \ref{tab:unibridge_focus_xlmr} and Table \ref{tab:unibridge_focus_mbert}.

\begin{table*}
    \centering
    \resizebox{0.9\textwidth}{!}{
    \begin{tabular}{l|cccccccccccccc}
          & \textbf{am} & \textbf{ang} & \textbf{cdo} & \textbf{crh} & \textbf{eml} & \textbf{frr} & \textbf{khm} & \textbf{kan} & \textbf{lij} & \textbf{pbt} & \textbf{san} & \textbf{snd} & \textbf{sin} & \textbf{som} \\
         \hline
         \hline
         FOCUS (XLM-R) & 45.72 & 40.13 & \textbf{43.03} & \textbf{46.03} & \textbf{41.53} & 45.24 & 35.12 & 45.85 & 40.09 & 43.24 & \textbf{30.15} & 50.67 & 51.22 & 46.89 \\ 
         UniBridge (XLM-R) & \textbf{49.6} & \textbf{43.24} & 42.91  & 46.03 & 40.15 & \textbf{50.67} & \textbf{42.67} & \textbf{48.72} & \textbf{45.16} & \textbf{46.09} & 29.74 & \textbf{51.32} & \textbf{52.86} & \textbf{54.22} \\ 
    \end{tabular}}
    \caption{FOCUS initialization and UniBridge with XLM-R backbone on WikiANN.}
    \label{tab:unibridge_focus_xlmr}
\end{table*}

\begin{table*}
    \centering
    \resizebox{0.9\textwidth}{!}{
    \begin{tabular}{l|cccccccccccccc}
          & \textbf{am} & \textbf{ang} & \textbf{cdo} & \textbf{crh} & \textbf{eml} & \textbf{frr} & \textbf{khm} & \textbf{kan} & \textbf{lij} & \textbf{pbt} & \textbf{san} & \textbf{snd} & \textbf{sin} & \textbf{som} \\
         \hline
         \hline
         FOCUS (mBERT) &\textbf{17.85} & 48.82 & 20.25 & \textbf{46.05} & \textbf{45.01} & \textbf{55} 15.83 & 43.16 & 43.17 & 36.55 & \textbf{21.24} & 40.83 & 11.07 & 55.85 \\ 
         UniBridge (mBERT) & 15.46 & \textbf{53.28} & \textbf{30.42} & 45.67 & 36.15 & 54.72 & \textbf{19.49} & \textbf{44.07} & \textbf{45.49} & \textbf{39.33} & 20.55 & \textbf{42.36} & \textbf{13.68} & \textbf{62.28}\\ 
    \end{tabular}}
    \caption{FOCUS initialization and UniBridge with mBERT backbone on WikiANN.}
    \label{tab:unibridge_focus_mbert}
\end{table*}

UniBridge surpasses FOCUS in performance across 10 out of 14 languages and 9 out of 14 languages on WikiANN. Among these languages, approximately 10-15\% of the tokens exhibit semantic alignment. We theorize that UniBridge's advantage lies in its ability to leverage these aligned tokens, which facilitates a smoother and quicker convergence during the subsequent MLM training phase compared to FOCUS initialization.

\subsection{UniBridge with Large Language Models}
\label{sec:mgpt-mbart-llms}

To evaluate UniBridge with large (decoder-style) Language Models (LLMs), we extended our experiments to include mGPT and mBART, alongside XLM-R and mBERT. This broader assessment demonstrates UniBridge's versatility and effectiveness across different model types. The results, presented in Table \ref{tab:mgpt-mbart-results}, highlight the strengths of UniBridge in diverse settings.

\begin{table*}[t]
    \centering
    \resizebox{0.9\textwidth}{!}{
    \begin{tabular}{l|cccccccccccccc|c}
          & \textbf{amh} & \textbf{ang} & \textbf{cdo} & \textbf{crh} & \textbf{eml} & \textbf{frr} & \textbf{khm} & \textbf{kan} & \textbf{lij} & \textbf{pbt} & \textbf{san} & \textbf{snd} & \textbf{sin} & \textbf{som} & \textbf{Average} \\
         \hline
         mGPT & 7.49 & 18.29 & 17.44 & 9.41 & 9.79 & 5.14 & 7.85 & 7.28 & 14.14 & 6.35 & 18.53 & 11.28 & 12.69 & 18.66 & 11.74 \\
         MAD-X (mGPT) & \textbf{63.1} & 51.15 & \underline{62.28} & 55.6 & \underline{43.79} & \underline{60.55} & 60.32 & \textbf{55.75} & \textbf{63.11} & 50.03 & \underline{61.62} & \textbf{56.66} & \underline{64.27} & \underline{61.36} & \underline{57.83} \\
         UniBridge (mGPT) & \underline{61.09} & \textbf{60.32} & \textbf{65.13} & \textbf{63.73} & \textbf{54.06} & \textbf{69.43} & \textbf{62.35} & \underline{55.38} & \underline{62.24} & \textbf{54.28} & \textbf{66.07} & \underline{54.51} & \textbf{66.42} & \textbf{70.29} & \textbf{61.81} \\
         \hline
         mBART & 19.19& 15.47& 10.46& 9.1& 14.92& 18.86& 13.16& 15.52& 6.22& 11.45& 19.31& 16.68& 13.21& 14.04& 14.11\\
         MAD-X (mBART) & \underline{67.03}& 51.24& 56.57& 29.73& 39.13& 51.5& 28.79& \underline{43.52}& \textbf{49.72}& \underline{45.25}& 51.85& \underline{58.64}& \textbf{60.33}& 51.2& 48.89\\
         UniBridge (mBART) & \textbf{69.15}& \textbf{67.5}& \textbf{67.89}& \underline{61.91}& \textbf{67.14}& \textbf{57.07}& \textbf{41.74}& \textbf{48.37}& \underline{44.1}& \textbf{52.47}& \textbf{60.99}& \textbf{59.12}& \underline{59.29}& \textbf{54.48}& \textbf{57.94}
    \end{tabular}}
    \caption{Various configurations with the mGPT and mBART backbone on the WikiANN dataset. We highlight in \textbf{bold} the highest F1 score and \underline{underline} the second highest of each target language for each backbone model.}
    \label{tab:mgpt-mbart-results}
\end{table*}

\end{document}